# Machine Learning Modeling to Evaluate the Value of Football Players

Chenyao Li, Dr Stylianos Kampakis, Prof Philip Treleaven

Department of Computer Science

University College London

## Abstract

*In most sports, especially football, most coaches and analysts search for key performance indicators using notational analysis. This method utilizes a statistical summary of events based on video footage and numerical records of goal scores. Unfortunately, this approach is now obsolete owing to the continuous evolutionary increase in technology that simplifies the analysis of more complex process variables through machine learning (ML). Machine learning, a form of artificial intelligence (AI), uses algorithms to detect meaningful patterns and define a structure based on positional data. This research investigates a new method to evaluate the value of current football players, based on establishing the machine learning models to investigate the relations among the various features of players, the salary of players, and the market value of players. The data of the football players used for this project is from several football websites. The data on the salary of football players will be the proxy for evaluating the value of players, and other features will be used to establish and train the ML model for predicting the suitable salary for the players. The motivation is to explore what are the relations between different features of football players and their salaries - how each feature affects their salaries, or which are the most important features to affect the salary? Although many standards can reflect the value of football players, the salary of the players is one of the most intuitive and crucial indexes, so this study will use the salary of players as the proxy to evaluate their value. Moreover, many features of players can affect the valuation of the football players, but the value of players is mainly decided by three types of factors: basic characteristics, performance on the court, and achievements at the club.*

## 1. INTRODUCTION

Football (Soccer) is the most popular sport in the World. It possesses the most estimated global fans in the world – nearly 4 billion fans around in nearly all regions. Therefore, due to many fans, football has a high commercial value which boosts it to become the highest-paid sport in the world (SPORTYTELL, 2021). Then as one of the highest-paid sports, it is very important to know the key parameters in deciding the salary of each player, because a rational salary structure is pivotal to increasing the revenue for football clubs. More importantly, the salary is also one of the most significant indexes to reflect the valuation of a player.

Evaluating the value of each player is indeed a necessary action because it will reflect both the comprehensive ability of players and the value of the players in the football market. Due to the transfer system in the football world, transfer fee means the actual prices that will be paid to the clubs, and the market value of a player will be one of the crucial references for football clubs to estimate the player's transfer fee. Although data analytic methods in the football world are still rarely used compared to other major sports there are still many different methods and models to estimate the value of the players (Kaplan, 2010). For instance, some methods will only consider the performances of players in the matches (e.g., goals, assists), and currently, some new methods will also consider the commercial value of the players (e.g., fans on Instagram).

Current football researchers are more likely to use some traditional linear regression models to explore the market value rather than some more complex ML models. However, linear regression models only assume the linear relation between a dependent variable and independent variables, but not all independent variables only have a linear relation with the dependent variable. So, this research will focus on two new ML models to measure the value of the players in the market. One is the normal multiple linear regression model while another model will use a completely new ensemble tree method known as the random forest model.

### 1.1. Scientific Contribution and Impact Statement

This study contributes and impacts to the science and football world in many ways stated below:
1) Present a new method to evaluate the value of players engaged in highly-commercialized sports.
2) Develop two new predictive models to predict the salary of the players by selected features of football players.
3) Evaluate the overestimation or underestimation of the football players.
4) Provide two models for football clubs to calculate the approximate salary given to the players, especially a completely new Random Forest model.
5) Explore the possible factors that may largely affect the value of the players by the models.

## 2. LITERATURE REVIEW

The factor about the player's achievements is the biggest difference between this research and previous ones. Most researchers incorporate the winning rate for all matches in a season as the standard to reflect the grade of the players (e.g., Muller, Simons, and Weinmann, 2017; Lucifora and Simmons, 2003). However, this research implements the final grades in different tournaments for the football clubs that players work for in that season as the factor to reflect the final achievements of the players. This approach is effective because the final grade in the tournaments for a season is more evident than just the winning rate for a season. Most importantly, according to the research of Deloitte Football Money League for the past 10 years[1], the research can easily get that the ranking for the clubs in different tournaments has a direct effect on the revenue of the clubs in that season which can better show the importance of final achievements in tournaments to the clubs.

---

[1] Link of Deloitte's report: https://www2.deloitte.com/uk/en/pages/sports-business-group/articles/deloitte-football-money-league.html

After the "Bosman judgement"[2], the transfer of players has become more convenient which caused many researchers to explore the method of evaluating the market value of the players. Because the market value of players directly reflects the comprehensive ability of players, it is pivotal to consider the true transfer value of players that football clubs will be paid.

Compared to predicting the true transfer fee of the players, the evaluation of players' market value considers more about the characteristics and performances of players themselves. In most research about market-value evaluation methods for football players (e.g., Muller, Simons, and Weinmann, 2017; Lucifora and Simmons, 2003; Frick, 2006; Franck and Nuesch, 2012), they all mainly utilize two types of factors for players to evaluate the market value of the players – basic characteristics of players (e.g., age, height) and the performance of the players (e.g., goals, assists). Nevertheless, some researches currently have focus on some other factors of the players, like the popularity of players (e.g., Muller, Simons, and Weinmann, 2017).

## 3. PLAYERS' EVALUATION AND MARKET ANALYSIS

### 3.1. Market Value of Players in Professional Football

In Herm's et al. research (Herm, Callsen-Bracker, & Kreis, 2014), they try to define what is the market value in the professional football world as "an estimate of the amount of money a club would be willing to pay to make [an] athlete sign a contract, independent of an actual transaction". Although the market value can be regarded as the most important proxy to evaluate the true transfer fee that may be paid by the club finally (He, Cachucho, & Knobbe, 2015), the market value of the players is still different from the true transfer fee. The evaluation of the market value for players depends on the information, characteristics, and performance of players themselves, and true transfer will consider some other complex factors, like other crucial indexes – the length of the current contract and some features of both buyer and seller's clubs (Trequattrini, Lombardi, & Nappo, 2012).

### 3.2. Relation Between Market Value of Players and Salary of Players

The research uses the salary of football players as the proxy to evaluate the market value of the players, so it is important to state why the salary of players can be regarded as a proxy to measure the value of the players. Frick's research (2006) has claimed that the determinations of the players' salary are player's age, the number of games that have played, the position of the players, and some other performance data of players, and other research about the players' market value (Müller, Simons, & Weinmann, 2017) has summarized the main factors that determine the player's market value: player characteristics (e.g. Age, Height), player performance (e.g. goals, assists), and player popularity (e.g. Wikipedia page views, Reddit posts). Both types of research have considered the basic characteristics and performance of players, so the factors that have determined the salary and value of the football players are highly matched for the conclusion of the two kinds of research.

---

[2] Bosman judgement is a very famous event about the transfer in the football world, and it provides huge convenience for the transfer of players.

### 3.3. Market Value/Salary – Player's Basic Characteristics Relation

Lucifora and Simmons's research (2003) had specific research on the relation between the age of football players and the earnings of the players. They concluded that normally soccer player's age has a positive impact on his salary but with diminishing rate, and 28 years-old age is a turning point for most football players, beyond this age, greater experience of players is not enough to offset the decline in physical conditions, but with the development of sports medical level, this turning point age may be older.

Bryson and Frick's (2009) researched how footedness[3] affects the salary of soccer players, which extends to the effect of different positions on the salary of players. Moreover, He, Cachucho, and Knobbe'e research (2015) have done more specific research about the effect of positions – they used the statistical method to prove that different positions of football players do affect the market value of players. Furthermore, another conclusion from their research was the classification of the position of players, normally, the position of football can be classified into 12 categories, but they have demonstrated that to measure the market value of players, it is also feasible to simplify and classify the football players into 4 groups – goalkeepers, defenders, midfielders, and strikes. Therefore, this research will utilize their way to classify the positions of players.

Table 2.1 has displayed the indicator of basic characteristics of players:

**Table 2.1: Player basic characteristics indicator to measure market value.**

| Player basic characteristics | Description |
|---|---|
| Age | The age of players reflects the physical condition and experience of players |
| Position | The position of players and divided into four categories of positions: <br> 1. Goalkeepers <br> 2. Defenders <br> 3. Midfielders <br> 4. Strikes |

### 3.4. Market Value/Salary – Player's Performance Relation

The performance data of players in the matches directly demonstrates the ability of players, and this type of data suits nearly all sports not only football, so all research about evaluating the market value of players will include the performance factors. According to combining many academic kinds of research about football player valuation, Muller, Simons, and Weinmann's research (2017) has summarized the factors that determine the market value of football players. In their research, player's performance will mainly divide into 8 categories, *Playing time; Goal; Assist* is the most used factor for nearly all research to measure the market value and salary of players, and the rest of the factors are not commonly used by some researches like passing and dribbling. This research includes the indicators from Muller's research except for indicators about *dribbles* and *fouls* (all shown in Table 2.2).

---
[3] Footedness: The natural preference for the left or the right foot.

Moreover, Franck and Nuesch's research (2012) have shown that some features shooting of players also affects the market value of football players, especially the number of shots on target which increases the market value of players by 95% quantile in their model. Therefore, the shooting factor of players is also considered in this research, and the number of shots on target will be one of the shooting predictors in the models, which this research will try to test the authenticity of Franck and Nuesch's results.

**Table 2.2: Player performance indicator to measure the market value**

| Player Performance | Description |
| --- | --- |
| Playing time | Playing time (mins or games) for players |
| Goal | Goals |
| Assists | Assists (help other players score a goal) |
| Passing | The number of passes to other players and the accuracy of passing |
| Shooting | The number of shots on target and goal/shot |
| Possession | The number of dribbles the ball, carry the ball and receive the ball |
| Defensive Actions | The number of a player's tackles, clearance, blocks, and interceptions |
| Cards | The number of yellow/red received by a player |

## 4. DATA SELECTION

### 4.1. Data Pre-Processing

This section will briefly state the process and results of the pre-processing of data obtained from two football websites, and how these data will be used in two studies to build the regression model. The main programming tool used for data pre-processing is the *Pandas and NumPy* library of python – which transfer all the data in Excel to the data frame to do further operation and processing.

### 4.2. Table Combining

1. After the research acquires the data, both Performance data from the website FBREF and Salary data from the website CAPOLOGY are separated into different leagues. The first step is to combine all the data from different leagues into one data table.
2. Performance data from the website FBREF for the players are separated into several tables, and each table corresponds to one type of character of players – including standard stats, shooting, passing, defensive actions, and possessions. Therefore, the study combines all feature tables for one season into an integrated table based on the name of the players and does the same operations for another two seasons.

## 4.3. Data Cleaning

After integrating all the character tables, all players will correspond to the data of all their features. But not all the features in the table are needed in the study, according to the research on the features of players, only some selected features' data will be used to build the model (all shown in Table 3.1). Thus, the first step is to extract the needed features' data from the table. Moreover, some players do not play much time in a season or even they never have a chance to play in a season, therefore, to keep the efficiency and accuracy of the data set, the research will only consider the players whose total playing time is larger than 90 minutes[4], and delete all the players who do not reach the 90 minutes requirements.

## 4.4. Data Type Transferring for some features

The salary data from the website CAPOLOGY also include the player's name, current club, and current league of the players. But these three factors are the string type, and they cannot directly be used to build the statistical model. So, the study needs to transfer these characters into the integer, and players with the same clubs or leagues will have the integer for those characters. The specific transfer operations are shown in Table 3.2:

**Table 3.2: Data type transfer table.**

| Position | League | Club |
|---|---|---|
| **Defenders -> 1** | Premier League -> 1 | AC Milan -> 1 |
| **Midfields -> 2** | Bundesliga -> 2 | Alaves -> 2 |
| **Strikes -> 3** | La Liga -> 3 | Angers -> 3 |
| | League 1 -> 4 | Arminia Bielefeld -> 4 |
| | Series A -> 5 | Arsenal -> 5 |
| | | Aston villa -> 6 |
| | | Atlanta -> 7 |
| | | …… |

Note:
1. There are 3 positions, 5 leagues, and 98 clubs in the research.
2. The value assigned to the club is based on the alphabetic order.

## 4.5. The calculation of player's total achievements

The study calculates a total achievement index from the ranking of three tournaments that this thesis has done the analysis and explanation in Section 3.1.3. According to the Section 3.1.3, this research has claimed that the domestic league, domestic league cup, and UEFA Champions League (UCL) will be counted in the total achievements index, but the domestic league and UEFA Champions League (UCL) are much more important than the domestic league cup, so domestic leagues and UCL will also have much more weights than domestic league cup. Then this study only counts 1/10 proportion for the domestic league cup, the rest proportion is equally distributed by domestic leagues and UCL (each 4.5/10 weight proportion).

---
[4] 90 minutes is normally the time of one complete football match.

Moreover, the value of the ranking also needs to do some transfer, because the smaller value of ranking means a better grade for clubs that have achieved it. Therefore, this study will get the reciprocal of their true ranking value to calculate their total achievements. The equation will be:

$$Total\ Grade = \frac{1}{League_{Rank}} * 4.5 + \frac{1}{UCL_{Rank}} * 4.5 + \frac{1}{League\ Cup_{Rank}} * 1$$

Therefore, the total points for this marking system for players' achievement are 10 points, and 10 points can only be achieved when the players get champions in all three tournaments.

## 5. MODEL SELECTION

### 5.1. Predictive Modelling and Analysis

The mixed-effect model is suitable for the data with several nested levels, and the data of this study is exactly hierarchical structure – the players are nested with their clubs and clubs are nested with their leagues (Figure 4.1).

Figure 4.1: The hierarchical level of players.

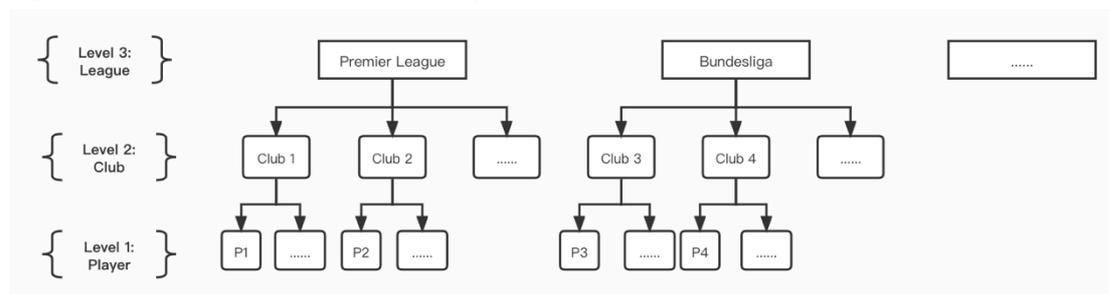

From Figure 2, level 1 is players, and players are playing/working for their clubs, so level 2 is the football clubs, and the clubs selected by this research are all belonging to the top 5 leagues. Theoretically, level one is the microlevel which is normally the individual factors, and the higher levels are the macrolevels which are normally the classification factors. Therefore, the hierarchical structure is reasonable for this research.

Nevertheless, this research is not interested in exploring a lot about the effect of specific clubs and leagues on the salary of players and mainly hopes to explore the relation between players' salaries and features other than clubs and leagues. But these two nested factors have some effects on the final predicted salary.

Therefore, this research will regard the *Player[5], Club and League* as the random effects, and other features in Table 3.1 will be the fixed effects. The formula is as follows:

$$\begin{aligned}Weekly\ Salary &= \alpha_{i(c(l))} + \beta * Player\ Basic\ Characters_{i(c(l))} + \gamma \\ &\quad * Player\ Perfomance_{i(c(l))} + \delta * Player\ Achievement_{i(c(l))} + \mu_{i(c(l))} \\ &\quad + \mu_{c(l)} + \mu_l + \varepsilon_{i(c(l))}\end{aligned}$$

$i(c(l))$ represents the index_i of a player who is nested with their clubs $c$ and further nested with their leagues $l$. $\alpha_{i(c(l))}$ is the individual intercept; $Player\ Basic\ Characters_{i(c(l))}$ include the factors of *Age* and *Position*; $Player\ Perfomance_{i(c(l))}$ consists of starts, total playing time, goal, assist, yellow card, red card, shot on target, goal per shot, total pass attempt, pass successful percentage, tackle, block, interception, clearance, total dribble attempt, dribble successful percentage, carry, being passed as the target, and successful percentage of receiving the ball; $Player\ Achievement_{i(c(l))}$ only includes the total grade value which is a marking system combining all rankings of each tournament that players have acquired in a season. $\mu_{i(c(l))}$, $\mu_{c(l)}$, $\mu_l$ are the random factors: $\mu_{i(c(l))}$ is the weekly salary observed for the same player $i$; $\mu_{c(l)}$ is the weekly salary observed for the players in the same football club; $\mu_l$ is the weekly salary observed for the clubs in the same league? $\varepsilon_{i(c(l))}$ is the remaining error. Moreover, the study will assume the random factors and error terms are independently and identically distributed.

## 5.2. Build Linear Mixed-effect Model

This research uses the 'smartest' (or 'lmer4') package of R language to build the model. The league and club will be regarded as random factors, and the research will set the club and league of players as the nested group. Figure 4.2 has shown the main codes of building the linear mixed effect model, and WEEKLY_GROSS is the dependent variable; all factors before (1|*League_num/Club_num*) are the fixed effects; (1|*League_num/Club_num*) is the random-effect term.

*(1 | g1/g2)* means the intercept among g1 and g2 within g1, and

$$(1\ |\ g1/g2) = (1\ |\ g1) + (1\ |\ g1{:}g2)$$

(Bates, Mächler, Bolker, & Walker, 2015)

Because the dataset of the study is grouped by the Club *(Club_num)* which is nested within League *(League_num)*, then (1 | *League_num/Club_num*) can better model the variation in the intercept and reflect the hierarchical structure of the dataset (Bates, Mächler, Bolker, & Walker, 2015).

Figure 4.2: Code of mixed-effect modelling.

---

[5] The study applies the average calculation for three seasons' data, and every cluster grouped by the player only has one player, so it is no need to set the player as the random effect for this study. But each player belongs to different clubs and leagues.

```
model <- lmer(WEEKLY_GROSS ~ Current_Age + POS + grade_value + Starts
              + Min + Gls + Ast + CrdY + CrdR + SoT + G_Sh
              + Pass_Att + Cmp_per + TklW + Blocks + Int + Clr
              + Dribble_Att + Dribble_Succ_per + Carries + Targ + Rec_per
              + (1 | League_num/Club_num),
              data=dataset, REML=TRUE)
```

### 5.3. Predictive Modelling of Salary through a Random Forest Model

Most past research about the market value uses the traditional linear regression model, but currently many other ML models can also handle the regression problems, so this study will apply another ML algorithm to build this salary predictive regression. The study has tried the full connected neural networks and random forests algorithms, and finally, this study chooses the random forest algorithms. The reason why the random forest algorithm is a suitable ML method for this study will be explained and discussed in the main section.

### 5.4. Model Selection – Random Forest Algorithms

Random Forest is a typical random tree ensemble method, and the random forest will combine several decision trees to get a better result. Moreover, the basic theory of random forest is the extension of bagging algorithms.

#### 5.4.1. Bagging Algorithms (Bootstrap Aggregation) and Random Forest (RF)

The bagging algorithms normally will be used to reduce the errors and variance of decision trees (Nagpal, 2017). Normally for the ensembled method, the combination of independent base estimators will have better performance and large reduce the errors, so to make the study more accurate, the study would better select estimators that are independent of each other. Therefore, if the study obtains a huge number of datasets without overlapping data, the model or learner normally can get a better conclusion. However, sometimes people cannot collect such huge numbers of data, and if researchers continue to use non-overlapped data for the separated estimator, then the researcher will only gain a small and unrepresentative sub-dataset which causes a bad performance for the model.

Therefore, bagging algorithms produce the different base estimators by bootstrap distribution (Zhou, 2012). The specific idea is to generate a sample containing the *m* training examples from the training sample randomly with replacement, and more importantly, the training examples can be selected repeatedly. By doing the generating process T times, the *T* subset of the data will be used to train each base estimator (decision trees for RF). Then bagging algorithms will ensemble different base estimators – the specific strategy is averaging for regression and voting for classification, and this study is to do regression research, so the research will apply the first strategy. Figure 4.3 (Zhou, 2012, p.49) can summarize the basic concepts of bagging algorithms above well.

```
Input: Data set D = {(x₁, y₁), (x₂, y₂), ..., (xₘ, yₘ)};
       Base learning algorithm 𝔏;
       Number of base learners T.
Process:
1. for t = 1, ..., T:
2.    hₜ = 𝔏(D, 𝒟_bs)    % 𝒟_bs is the bootstrap distribution
3. end
Output: H(x) = arg max_{y∈𝒴} Σ_{t=1}^{T} 𝕀(hₜ(x) = y)
```

Figure 4.3: Basic idea of the Bagging Algorithm

Note: Zhou, 2012, p.49

The random forest contains several base estimators to combine their results in some strategies mentioned in bagging algorithms – averaging for regression and voting for classification – to get better performance. So Random forest is a typical extensive method of the bagging algorithms, but RF has one big difference from the normal bagging algorithms – RF will take random selection for features (predictors in the research, like age, goal……) rather than always using all features to train the decision trees.

## 5.5. Decision Tree and Random Forest

A decision tree is a tool that uses a set of tree-structured decision tests working in a divide-and-conquer way (Zhou, 2012). All non-leaf nodes are called the splits which represent the decision rule or feature test and based on the different results from the feature test, data will fall to the nodes at the next level and split into different subsets. Each leaf node displays the outcome of the prediction.

Decision trees can be used for solving both regression and classification problems and two types of decision trees also use the different decision rules. Moreover, different decision trees even have different structures, like ID3 (polytree structure), C4.5 (polytree structure), and CART (binary-tree structure). This study utilizes Python's sci-kit-learn package to do the training and testing for the model, and the decision tree used by the scikit-learn package is the **classification and regression tree (CART)**. Therefore, the explanation of this thesis will only pay attention to the process of CART to handle the regression problem, and the thesis will concisely explain the working process for one **regression tree** of CART below:

Given the datasets: $D = \{(x_1, y_1), (x_2, y_2), ..., (x_N, y_N)\}$, and one formula can best display the relation between outcome y and features x (one or more):

$$\hat{y} = \hat{f}(x) = \sum_{m}^{M} c_m I\{x \in R_m\} \quad (\boldsymbol{Equation\ 4.1})$$

Each instance finally will drop to one leaf node (a subset of $R_m$: $R_j$) and get a predicted income $\hat{y} = c_l$, and $c_l$ is the average value of the output of all training samples in the leaf node $R_j$. The specific process of growing a regression tree is:

**Step 1:** Sort the data based on $x_j$, then find the value for one of the predictors $x_j$ to split the data into 2 parts and define two ranges (two nodes on the tree):

$$R_1(j,s) = \{x \mid x_j > s\} \text{ and } R_2(j,s) = \{x \mid x_j \leq s\}$$

**Step 2:** This step is to choose the best value of step 1. To choose the best feature of $j$ and the best splitting point of $s$ for each feature, the model needs to iterate all the features $j$ at first and then iterate all the splitting values s to get the best $(j, s)$. To reach this target, the trees will use the Mean Squared Error (MSE[6]) as the standard to evaluate the accuracy of the model – the smaller the MSE, the better the value of $s$. So, the suitable value of $s$ is the minimum MSE for nodes, the formula can be written as:

$$\min_{j,s}[\min_{c_1} \sum_{x_i \in R_1(j,s)} (y_i - c_1)^2 + \min_{c_2} \sum_{x_i \in R_2(j,s)} (y_i - c_2)^2]$$

$$Note: c_1 = average(y_i \mid x_i \in R_1(j,s)), c_2 = average(y_i \mid x_i \in R_2(j,s))$$

**Step 3:** Put the best $(j, s)$ in the tree structure to split the dataset with feature $j$ and splitting value $s$, then get the prediction:

$$\widehat{c_m} = \frac{1}{N_m} \sum_{x_i \in R_m(j,s)} y_i \; ; \quad x \in R_m, m = 1,2$$

**Step 4:** Continue to execute Step 1-3 for the sub-ranges/nodes ($R_1$ and $R_2$) produced in the last iteration until meeting the stop condition (like only one training sample at all nodes).

**Step 5:** Until Step 4, a decision tree will be produced, and this kind of decision tree is normally called the **least-squares regression tree**. If there are M subsets of leaf nodes ($R_1, R_2, ..., R_m$) and each subset gets a prediction $c_m$, then the comprehensive equation of regression trees can be expressed as **Equation 4.1**.

**The random forest** contains several regression trees explained above and combines all the results of each tree to calculate the average of these results which will be the final prediction of the model. (There is a part of one regression tree concluded from the model that displays in the next chapter in Section 5.2)

## 5.6. Model Analysis

Random forest will firstly randomly select a sample containing m training examples from the total training sample with replacement and the training examples of each sample can be selected repeatedly. Therefore, the RF algorithm can also be used for research that has not collected a huge number of data. For this study, due to the limited number of players in the top 5 leagues and the pre-processing of the training datasets, there are only 1684 players' data that can be used for training the model. ML methods like neural networks will be more accurate by training with a large training dataset and compared to neural networks, RF will be a more efficient method to utilize the training datasets.

---

[6] $MSE = \frac{1}{n}(\sum_{i=1}^{n}(y_i - \hat{y}_i)^2)$, $y_i$ is the true value and $\hat{y}_i$ is the prediction.

Moreover, the RF algorithm is suitable for handling non-linear data or categorical features. Because RF will always iterate to choose the best splitting point for a feature and then for the value, this splitting way is functional for both categorical values and continuous values. The dataset for this research has three features which are set as the categorical values – Position, League, and Club.

RF algorithms also work well for high dimensionality data and even train faster than some other ML methods. RF will randomly separate the total training dataset into several subsets and randomly choose the number of training samples and features, which means each subset will handle part of the features, so the model can easily handle the dataset with many features even hundreds of features. This is also why the RF may have a faster training speed – each decision tree will only handle part of the features in training datasets, but the neural network will always work for all features in every training epoch. To explore the effect of football players' features more concisely on their salary and market value, the study selects 24 features which is much more than other football researchers, so RF is indeed suitable for this dataset.

More importantly, for most ML models, if these models hope to get better fitting and improve the accuracy of the model, the researchers need to do the normalization for the training datasets. But RF can completely ignore this step, because the normalization of the training dataset is unable to influence the choice of the best splitting feature and value for trees, and hence the normalization of data is meaningless for RF. The dataset includes 24 features of the players and all of them have different ranges, so the values and ranges of each feature have a huge difference. Then RF can make me ignore the normalization step and keep the initial dimension of datasets, and the target is to keep the initial dimension of the values for each feature because this model will be easily used for other people – only input the data of players and no need to do more pre-processing for the datasets.

### 5.7. Build a Random Forest model

This study utilizes the scikit-learn package of Python to build the model. The scikit-learn package provides the method "*RandomForestRegressor*" to build the RF algorithms based on this research's processed training datasets. The following figure is to display the code for building the RF model:
Figure 4.4: Code of random forest modelling.

```python
rf = RandomForestRegressor(max_features=None,n_estimators=200,max_depth=None,
                           min_samples_split=2,min_samples_leaf=1,
                           criterion="squared_error",random_state=2)
```

Some key parameters of this method need to be set, and some values of parameters are acquired by doing the hyperparameter optimization based on the "*GridSearchCV*" method in the scikit-learn package (Codes on GitHub). Then the thesis has shown the meaning of all key parameters and their setting values in Table 4.1:

**Table 4.1: The meaning of variables in method "*RandomForestRegressor*".**

| Variable | Meaning |
|---|---|
| max_features | The maximum features that the tree will be selected from, and set to 'None' mean the number of features selected by the tree will always be smaller or equal to the total features of the training dataset. |
| n_estimators | The number of base estimators (regression trees) means this RF model will combine the results from 200 regression trees. |
| max_depth | The maximum depth of the regression trees and 'None' means the tree will grow without limits of depth. |
| min_samples_split | The minimum number of training samples that required to split the internal nodes. |
| min_samples_leaf | The minimum number of training samples at the leaf nodes, so this RF model will split to only one training sample for all leaf nodes. |
| criterion | The method that uses for measuring the quality of a split. For the regression tree, MSE is commonly used. |
| random_state | Control the randomness and bootstrapping of selecting the samples from the total training dataset. |

Therefore, based on the explanation above, the step of building the RF model for this study can be displayed in the following figure:

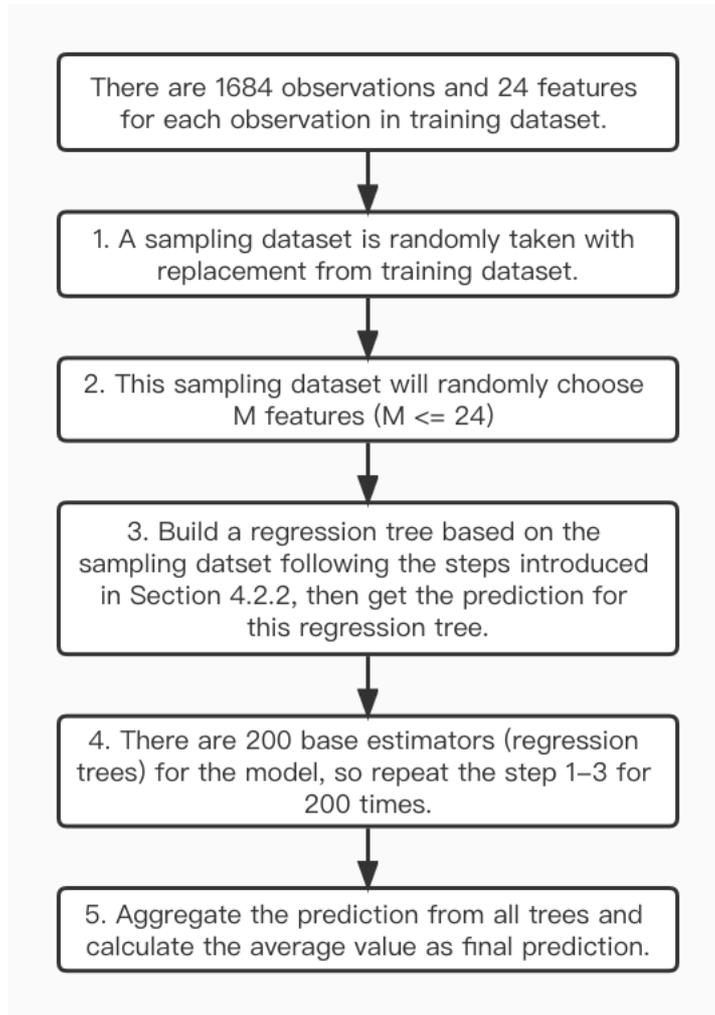

Figure 4.5: Specific process of building the RF model.

# 6. TESTING RESULT

**5.1: Predictive Modelling of Salary through a Multiple Linear Regression Model**

Model 1 - Salary ~ Basic Characteristics

**Table 5.1: Result of model 1.**

| Dependent Variable: Weekly Salary | | | | |
|---|---|---|---|---|
| **Fixed effects:** | | | | |
|  | Estimate | t value | Pr(>|t|) | |
| (Intercept) | -87535.7 | -6.523 | 4.99E-07 | *** |
| Current Age | 4214.13 | 11.337 | < 2e-16 | *** |
| POS | 13178.41 | 7.71 | 2.19E-14 | *** |

*Note: '*' is p < 0.05; '**' is p < 0.01; '***' is p < 0.001.*

*Number of observations is 1684 players. Number of groups: Players -> 1684; Club -> 98; League -> 5*

Model 1 is the primary model and only includes the *intercept, current age of players, and position of players*. **Pr(>|t|)** represents the **p-value** of each feature which display the statistical significance of each predictor in the model, and if the p-value is smaller, the statistical significance of the observed difference will be greater (Beers, 2022). Normally, if p is smaller than 0.05 for a feature, this feature has a fairly strong relationship with the response variable (Beers, 2022). In Model 1, both age (p < .001) and position (p < .001) of players are significant factors to determine the salary of players.

Model 2 - Salary ~ Basic Characteristics + Player's Performance

**Table 5.2: Result of model 2.**

**DEPENDENT VARIABLE: WEEKLY SALARY**

| FIXED EFFECTS: | | | | |
|---|---|---|---|---|
| | Estimate[7] | t value | Pr(>|t|) | |
| **(INTERCEPT)** | -40474.327 | -2.23 | 0.02814 | * |
| **CURRENT_AGE** | 2312.503 | 6.707 | 2.75E-11 | *** |
| **POS** | -141.831 | -0.056 | 0.95504 | |
| **STARTS** | -2833.973 | -2.451 | 0.01436 | * |
| **MIN** | -15.992 | -1.1 | 0.27138 | |
| **GLS** | 7089.504 | 7.76 | 1.51E-14 | *** |
| **AST** | 1276.876 | 1.407 | 0.15965 | |
| **CRDY** | 746.201 | 1.082 | 0.27937 | |
| **CRDR** | 3682.799 | 0.838 | 0.40197 | |
| **SOT** | 2249.046 | 5.058 | 4.72E-07 | *** |
| **G_SH** | -47773.578 | -2.822 | 0.00483 | ** |
| **PASS_ATT** | 50.414 | 2.901 | 0.00378 | ** |
| **CMP_PER** | 84.194 | 0.426 | 0.67002 | |
| **TKLW** | 17.347 | 0.096 | 0.92376 | |
| **BLOCKS** | -158.761 | -0.905 | 0.36537 | |
| **INT** | -253.544 | -1.247 | 0.21273 | |
| **CLR** | 176.575 | 3.15 | 0.00166 | ** |
| **DRIBBLE_ATT** | 131.946 | 1.493 | 0.13568 | |
| **DRIBBLE_SUCC_PER** | 66.84 | 0.821 | 0.41182 | |
| **CARRIES** | 118.389 | 4.618 | 4.18E-06 | *** |
| **TARG** | -78.829 | -4.576 | 5.11E-06 | *** |
| **REC_PER** | 5.578 | 0.031 | 0.97498 | |

*Note: '*' is p < 0.05; '**' is p < 0.01; '***' is p < 0.001.*

*Number of observations is 1684 players. Number of groups: Players -> 1684; Club -> 98; League -> 5*

Model 2 adds the factors about players' performance. In Model 2, the *age (p < .001), goals (p < - .001), shots on target ('SoT', p < .001), the number of Carries (p < .001), and the success rate of receiving the* pass from teammates ('Rec_per', p < .001) are most significantly related to the salary of players. Moreover, *the goals per shot ('G_Sh', p < .01), the number of passes ('Pass_Att', p < .01), Clearance (p < .01), and the starts of players in the matches ('Start', p < .05)* also can be regarded as the important factors to affect the salary of the players.

---
[7] **Estimate** represents the specific linear effect of each predictor to response variable – if one predictor is changed and other predictors are not changed, how the response variable will change when the changed predictor increase 1 unit.

Model 3 - Salary ~ Basic Characteristics + Player's Performance + Player's Achievement

**Table 5.3: Result of model 3.**

**DEPENDENT VARIABLE: WEEKLY SALARY**

| FIXED EFFECTS: | | | | |
|---|---|---|---|---|
|  | Estimate | t value | Pr(>|t|) | |
| **(INTERCEPT)** | -48173.23 | -2.685 | 0.00839 | ** |
| **CURRENT_AGE** | 2281.23 | 6.653 | 3.93E-11 | *** |
| **POS** | 232.77 | 0.093 | 0.92597 | |
| **STARTS** | -2494.11 | -2.163 | 0.03071 | * |
| **MIN** | -15.56 | -1.076 | 0.28228 | |
| **GLS** | 6878.1 | 7.553 | 7.14E-14 | *** |
| **AST** | 832.99 | 0.918 | 0.35863 | |
| **CRDY** | 914.91 | 1.332 | 0.18307 | |
| **CRDR** | 5481.14 | 1.251 | 0.21103 | |
| **SOT** | 2192.14 | 4.95 | 8.20E-07 | *** |
| **G_SH** | -49197.52 | -2.918 | 0.00357 | ** |
| **PASS_ATT** | 51.59 | 2.985 | 0.00288 | ** |
| **CMP_PER** | -44.34 | -0.225 | 0.82231 | |
| **TKLW** | 49.84 | 0.276 | 0.78243 | |
| **BLOCKS** | -156.76 | -0.898 | 0.36929 | |
| **INT** | -312.33 | -1.541 | 0.12348 | |
| **CLR** | 182.7 | 3.271 | 0.00109 | ** |
| **DRIBBLE_ATT** | 187.43 | 2.118 | 0.03429 | * |
| **DRIBBLE_SUCC_PER** | 93.84 | 1.158 | 0.24724 | |
| **CARRIES** | 107.24 | 4.192 | 2.92E-05 | *** |
| **TARG** | -78.72 | -4.597 | 4.63E-06 | *** |
| **REC_PER** | 47.44 | 0.268 | 0.78836 | |
| **GRADE_VALUE** | 9791.35 | 6.67 | 3.82E-11 | *** |

Note: '*' is p < 0.05; '**' is p < 0.01; '***' is p < 0.001.

Number of observations is 1684 players. Number of groups: Players -> 1684; Club -> 98; League -> 5

Model 3 adds the factor about players' achievements. The model 3 is like the model 2, and the significance of factors that affect the salary of players is nearly the same as the model 3, except for the additive factor of *the achievement of the players ('grade_value', p < .001)* which is also the most statistically significant factors related with the salary of players.

## 6.1. Comprehensive Analysis for Results

According to the final model – model 3, research has concluded the statistically significant factors that affect the value of the salary of players. In most other people's research about evaluating the market value of the players, *Age; Goals; Pass; Dribbles* are indeed important factors to determine the market value of the players (e.g. Müller, Simons, & Weinmann, 2017). These features are all the direct factors to reflect the ability of players and other researchers have already discussed and analyzed a lot, so the result from this research's model for these features is reasonable, and the thesis still hopes to highlight that *Age* and *Goal* are two of most significant factors to affect the weekly salary of players. Moreover, in this study's model, when other factors are not changed, if the player can get one more goal, their weekly salary can increase by € 6878.1; if the players increase their one-year age, they can get € 2281.23 more weekly salary.

Shots on target are another most significant feature that will largely affect the salary of players, and the result of the study has perfectly proved Franck and Nuesch's research (2012) about the effect of shots on target on the market value of players.

However, there are still some differences between the conclusion of these researches and this study. Compared with Muller, Simons, and Weinmann's research (2017), the features of *Assist* and *total playing time ('Min')* do not have strong statistical relation with the salary of players based on the p-value. For the *Assist*, if a player gets one more assist, then he will get € 832.99 more weekly salary; but for the *total playing time ('Min')*, it even negatively affects the salary of players but a very slight change of salary – one-minute playing time increase will cause € 15.56 decrease of weekly salary. Furthermore, the result also displays some other features that have strong relationships with the salary of players and are not mentioned a lot by other research – *Clearance; Goal per shot; Carry; Target of passing; Starts*, and due to the difficulty of collection of these features, these factors are not investigated a lot by other researches. *Carry* and *Target of passing (p < .001)* are other two features that have the most significant effect on the weekly salary of players, and they both have a positive effect on the value of the salary for players; but *Goal per shot, Target of passing, and Starts* have the negative effect for the value of players' salary.

Another important result is to demonstrate the initial assumption of this study that the achievement of tournaments for players is also one of the most significant factors affecting the salary and market value of players. For the index of expressing the achievements of tournaments (*'grade_value'*), it is an easy marking system set by this study to combine the rankings of three tournaments for a season (Section 3.3) and calculate the average points, so the total points are 10 points. According to the model, if the player can increase 1 more point under this achievement marking system, their weekly salary can increase by € 9791.35.

## 6.2. Model Evaluation

In some other research (e.g. Müller, Simons, & Weinmann, 2017), market value is an unobservable and not a completely precise factor, so it is difficult to evaluate in their research, and these researchers normally will compare their predicted market value with the true transfer fee, but actually, the true transfer fee is more complicated to evaluate and determined by many features other than features that determine the market value (Have explained in Section 2.1.1). Compared to these researches, this research has a clear proxy to evaluate the market value of players – the true salary of the players. The true salary of players will use to do the training model, and after training, it can still be a proxy to compare with the predicted salary for evaluating the accuracy of the model and getting the conclusion of overestimation/underestimation of players (Display in Section 5.1.3).

The following three tables have displayed the performance of three models in Section 5.1.1:

**Table 5.4: Performance for model 1.**

| AIC | BIC | $R^2$(cond.) | $R^2$(marg.) | RMSE |
|---|---|---|---|---|
| 42028.235 | 42060.808 | 0.481 | 0.06 | 57961.433 |

**Table 5.5: Performance for model 2.**

| AIC | BIC | $R^2$(cond.) | $R^2$(marg.) | RMSE |
|---|---|---|---|---|
| 41144.162 | 41279.885 | 0.596 | 0.357 | 47688.437 |

**Table 5.6: Performance for model 3.**

| AIC | BIC | $R^2$(cond.) | $R^2$(marg.) | RMSE |
|---|---|---|---|---|
| 41090.253 | 41231.405 | 0.606 | 0.432 | 47591.728 |

Akaike Information Criterion (AIC) and Bayesian Information Criterion (BIC) are two statistical indexes to score the model based on the model's log-likelihood and complexity (Brownlee, 2019). Normally the values of AIC and BIC are smaller, and the goodness of fit will be better. For this study, three models can detect the goodness of fit for each model when more features adding to the model. In model 2, the AIC has dropped from 42028.235 to 41144.162, so compared with the model 1, the model 2 has better goodness of fit; and in model 3, the AIC has dropped from 41144.162 to 41090.253, then even though the model 3 only adds one factor, the model 3 has improved the goodness of fit compared to the model 2. If using the AIC to select the model, researchers just simply choose the model with the smallest AIC overall considered models (Hastie, Tibshirani, & Friedman, 2016), so in this research, model 3 with all three types of features are the best model.

RMSE is the root mean square error: $RMSE = \sqrt{MSE} = \sqrt{\frac{1}{n}\left(\sum_{i=1}^{n}(y_i - \widehat{y_i})^2\right)}$, and $y_i$ is the true value and $\widehat{y_i}$ is the prediction. RMSE is to measure the variance between the true value and predicted value, so if the RMSE is smaller, the estimation of the model will be more accurate. For three models, the RMSE of model 1 is 57961.433; the RMSE of model 2 reduces to 47688.437 compared to model 1; the RMSE of model 3 reduces to 47591.728 compared to model 2, then the estimation of model 2 is better than the model 1 and the estimation of model 3 is better than the model 2.

The $R^2$ is the R-square which is one of the most common coefficients to determine and evaluate the accuracy of the regression model. The R-square can be calculated as:

$$R^2 = 1 - \frac{RSS}{TSS}, RSS = \sum_{i=1}^{n}(y_i - \hat{y}_i)^2, TSS = \sum_{i=1}^{n}(y_i - \bar{y})^2$$

$y_i$ is true value in the sample; $\hat{y}_i$ is the predicted value; $\bar{y}$ is the mean value of the sample

The range of R-square is normally 0 to 1, and the value of R-square is more closed to 1, the model has a better fit and the outcome y can be better accounted for by features x (one or more).
$R^2$(marg.) is the marginal r-squared values which express the total variance only explained by the fixed effects; $R^2$(cond.) is the conditional r-squared values that express the total variance of both the fixed effects and random effects (Nakagawa, Johnson, & Schielzeth, 2017). For the final model of study, the $R^2$(cond.) is 0.606 which is not a bad result but also not an ideal result, then the study will still utilize this model to evaluate the market value of the players, but this result facilitates another study – to apply a different ML model to try to get more accurate predictions.

The study will put the training datasets into the trained model to get the predicted salary of all players, then the study compares the predicted salary and true salary of all players to plot the graphs of true salary VS predicted salary (Figure 5.1) and residual plot (Figure 5.2):

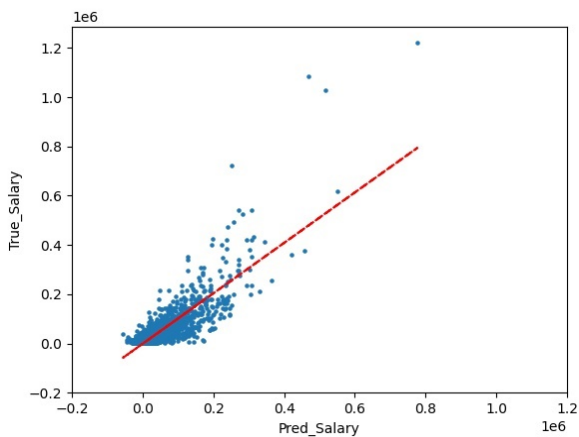
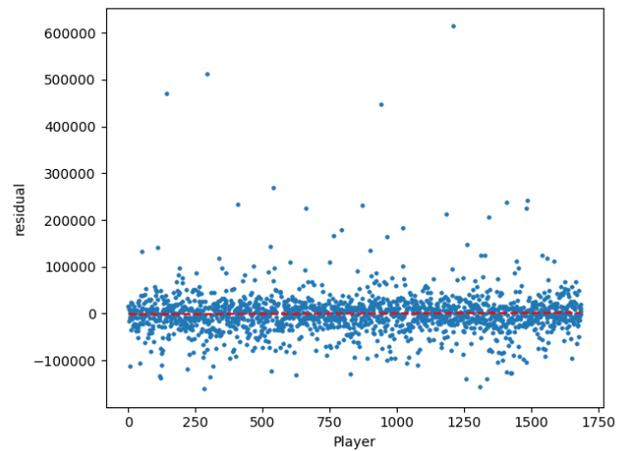

*Figure 5.1: Predicted Salary VS True Salary for a mixed-effect model.*    *Figure 5.2: Residual Plot for the mixed-effect model.*

In Figure 5.1, if the blue points are closer to the red line, then predicted salaries are closer to true salaries, and if the blue points are on the red line, it means predicted salaries equal to true salaries. From the plot in Figure 5.1, many predicted results are indeed not accurate, because many blue points are not close to red lines which means some players' predicted salaries have a large variance. Furthermore, due to the fit of the multiple linear regression model, some players' predicted salaries are even negative values, so this model is not significantly ideal, and another ML model will be applied in the next study.

Figure 5.2 has shown the residual[8] of each player in the dataset, which means plotting the error between an observed actual weekly salary for each player and a predicted weekly salary by this study's model (Gohar, 2020). To validate the residual plots, the study should determine the randomness and non-patterns of the residuals and inspect the normal distribution of the residuals (Gohar, 2020). Hence, this study firstly checks whether all residuals are randomly scattered around the line of zero, and the study should check that more residual points are closer to the line of zero. Therefore, the residuals in this study can be concluded as independent and normally distributed, and this study's regression model is valid.

## 6.3. Underestimation of Overestimation of Players' Value by Predictive Model

### 6.3.1. Method of Player's Value Estimation

Due to the value of R-square for the linear mixed effect model (0.606) being mediocre, the predicted salary for some players may exist a large bias, then if this study directly uses the predicted salary and true salary to do the comparison and conclude the overestimation or underestimation or normal estimation of players, the results will not be accurate. Therefore, to make the evaluation of market value more precise, this study calculates the prediction interval for each player's predicted value and uses the range of prediction intervals to judge the value of players.

The study utilizes the '*predict internal()*' function of the '*merTools*' package of R language to calculate the prediction interval for players' predicted values (Figure 5.3). To generate the prediction interval, the function will roughly follow the following steps (RDocumentation, n.d.):
1. This function firstly calculates the stimulated distribution of all parameters in the model.
2. From the multivariate normal distribution of fixed and random coefficients, separately take k draws.
3. Based on these draws from step 2, calculate the simulated values of linear predictor for each observation in the dataset for each simulation.
4. Acquire the upper and lower limits of simulated prediction values required by the function ('level' parameter).

Figure 5.3: Codes of calculating the prediction interval.

```
pred = predictInterval(merMod = model, newdata = test,
            level = 0.90, type="linear.prediction",
            include.resid.var = 0, seed = 10)
```

The study has required a 90% prediction interval for the model, then the predicted salary; upper limits of prediction interval; lower limits of prediction interval will be displayed in Figure 5.4:

Figure 5.4: Example of the prediction interval.

---

[8] Residual = Observer value – Predicted value

| Player | Salary | Pred | up | down |
|---|---|---|---|---|
| Aaron Con | 27212 | 10879.87 | 39224.15 | -16939.2 |
| Aaron Cre: | 58958 | 74213.18 | 102719 | 45082.38 |
| Aaron Hick | 11154 | 930.7504 | 27478.48 | -27175 |
| Aaron Ran | 172500 | 156346.2 | 181748 | 131187.2 |
| Aaron Wa | 106125 | 118190.8 | 147736.2 | 90516 |
| Abdon Pra | 4423 | 11811.56 | 38664.55 | -16117.4 |
| Abdou Dia | 103846 | 216888.6 | 244518.6 | 188808.6 |
| Abdoulaye | 6346 | 2335.175 | 29094.01 | -25413.9 |
| Abdoulaye | 88438 | 122814.1 | 148858.3 | 96752.18 |
| Abdoulaye | 88438 | 64679.87 | 90934.57 | 39702.52 |

According to the results of the prediction interval, this study can get the final analysis for the evaluation of the players. 90% prediction interval means that the value of the predicted salary of players will be in the prediction interval (values between the upper limit and lower limit) with 90% probability.

Therefore, if the true salary of a player is lower than the lower limit, this player will be judged as the **Underestimation**; and if the true salary of a player is higher than the upper limit, then this player will be judged as the **Overestimation**; and if the true salary of a player is between the upper limit and lower limit, then this player will be judged as the **Normal** estimation.

### 6.3.2. CART Trees

The fitting process of random forest regression algorithms is completely different from the traditional linear regression model because RF is a non-linear algorithm. Random Forest concludes several decision tree structures, so the testing dataset will input into several regression tree structures and average all the predicted values from each tree to get the final prediction.

The study has 200 regression tree structures, and the scikit-learn package provides the attribute '*estimator_[i]*' of method "*RandomForestRegressor*" to list the collection of fitted sub-estimators (sklearn.ensemble.RandomForestRegressor, n.d.). Then researchers can use the coding to draw the regression tree structure based on the values of estimators in the collection, and this study draw the first estimator as an example, but the tree structure is large, so the thesis only cut part of the tree to put into as an instance to show.
(Complete first regression tree plot is pushed on GitHub)

Figure 5.6: Part of one regression sub-tree in RF.

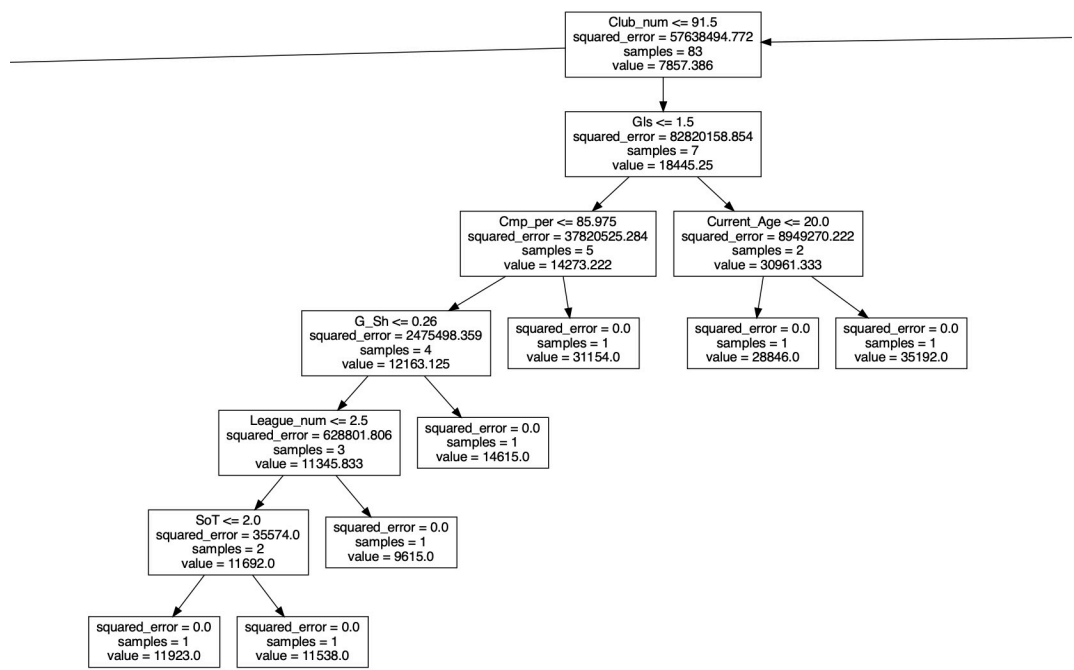

### 6.3.3. Feature Importance

The scikit-learn package also provides the attribute '*feature_importances_*' of method '*RandomForestRegressor*' to print the impurity-based feature importance, which is computed as that iterate all the splits and measure the total normalized reduction of the criteria (MSE) when the feature is used (sklearn.ensemble.RandomForestRegressor, n.d.). The value of this index is higher, the feature will be more important for this model. Table 5.7 has printed the values of each feature by coding. Then based on the value of feature importance, this study has also drawn the importance feature plot to make the results clearer.

**Table 5.7: Feature importance table.**

| Feature | Value | Feature | Value |
|---|---|---|---|
| **Current Age** | 0.030614 | **Success rate of pass** | 0.016148 |
| **Position** | 0.001907 | **Tackle** | 0.0204031 |
| **Achievement** | 0.208936 | **Blocks** | 0.064123 |
| **Starts** | 0.010796 | **Interception** | 0.0181336 |
| **Playing time** | 0.016453 | **Clearance** | 0.015099 |
| **Goal** | 0.057879 | **Dribble** | 0.118531 |
| **Assist** | 0.021266 | **Success rate of dribble** | 0.015121 |
| **Yellow card** | 0.012173 | **Carry** | 0.0262284 |
| **Red Card** | 0.004161 | **Pass as target** | 0.177547 |
| **Shots on target** | 0.074736 | **Success rate of receiving** | 0.0116589 |
| **Goal/Shot** | 0.014135 | **League** | 0.0242897 |
| **Pass** | 0.011364 | **Club** | 0.0282977 |

The value of the feature importance is larger, this feature is more important to determining players' salary, so according to the feature importance table, the study has ranked the importance of each feature in Figure 5.7:

Figure 5.7: Feature importance plot.

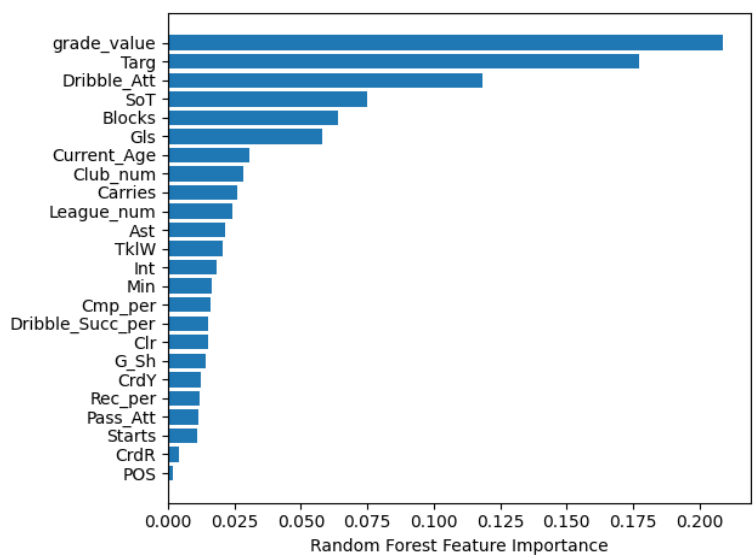

From the plot of feature importance, the achievements of players ('*grade_value*') are still one of the most important factors to affect the salary of players which is the same result as the mixed-effect model.

Some commonly researched factors – *Goal; Dribble; Age* – are still proved that they are the most crucial factors to affect the salary of players. Moreover, the important value of *Shots on target (SoT)* can also demonstrate the research of Franck and Nuesch (2012) – it is the key factor to determine the salary of players. More importantly, in the mixed-effect model, these features are also concluded to have a significant effect on the weekly salary of players. However, one difference is that *the number of blocks* plays an important role to decide the weekly salary in the RF model.

Compared with the conclusion of Muller, Simons, and Weinmann's research (2017), *Assist; total playing time ('Min')* also does not show strong significance with the salary of the players, which is also the same conclusion acquired from the mixed-effect model.

Other factors also have some effects to decide the salary of players but are not important, especially for the *position* and *red card ('CrdR')*, these features nearly do not affect determining the weekly salary of players.

## 6.4. Model Evaluation
### 6.4.1. Cross-Validation

After building the model and before confirming the best value of parameter inputting to build the model, the study firstly uses the cross-validation test to do evaluate the model with different parameters, so this testing is beneficial to do the hyperparameter optimization for the regression model (Combine with "*GridSearchCV*" method in scikit-learn package). There are several cross-validation test methods, but this study will only introduce the K-Fold Cross Validation because the study uses the "*cross_val_score()*" function in a scikit-learn package, and this function applies the K-Fold Cross-Validation. Figure 5.8 (Cross-validation: evaluating estimator performance, n.d.) has shown the thinking of the K-Fold Cross-Validation and the specific steps are:

1. Split the total dataset into k small subsets (k folds).
2. The model will be trained by k-1 subsets of total subsets
3. The model will be tested by remaining one data subset and get the scoring of the currently trained model
4. Change the testing dataset of k folds until every small subset is regarded as the testing dataset for one time.
5. Finally, there will be k scoring of the model, and the average value of k scoring is the final result of cross-validation.

### 6.4.2. R-Square Testing for Total Dataset

The study uses cross-validation and set R-square as the scoring method to find the model with the best parameters. However, like training and calculating the R-square for a mixed-effect model in the last study, this study also does not split the training dataset and testing dataset when training the model, and the study uses all observations in the dataset to do the training. After obtaining the trained RF model, the study just puts all the observations into the trained model to get the predicted weekly salary. The reason for not splitting the dataset is that the target of this research is to evaluate the market value of all players in the total dataset based on the predicted result of the model, and the study can get the most accurate salary prediction for all players in the dataset by using the trained model to predict the salaries for the training dataset because the model can always be highly-fitted with training dataset; another reason is that the research does not collect the huge number of data, so if this study split the dataset, the training data are not enough.

Then by getting the predicted salary for all players from the trained model and calculating the R-square, the R-square result of RF is: $R^2 = 0.948$.

This result is good, and the model is highly-fitted with a training dataset. More importantly, this model's result is further better than the result of the mixed-effect model ($R^2 = 0.606$), and the study also plots graphs of true salary VS predicted salary (Figure 5.10) and residual plot (Figure 5.11):

.

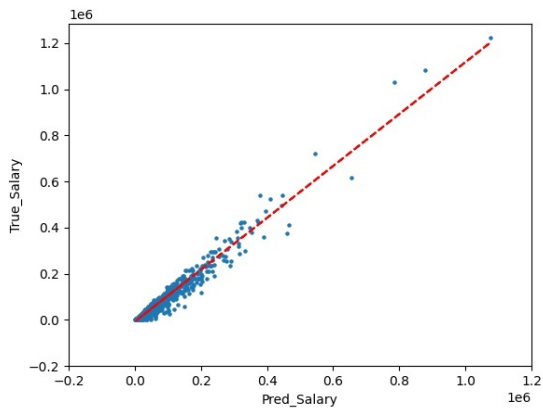
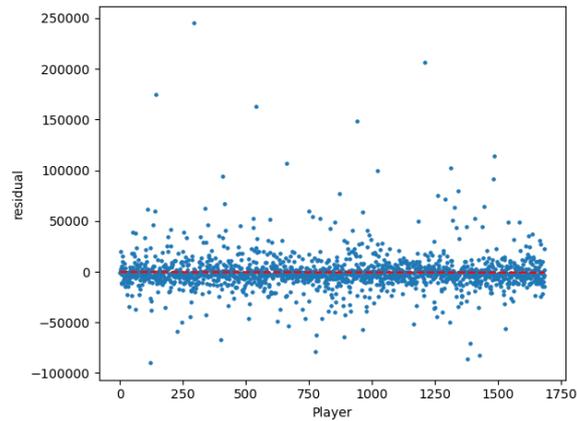

Figure 5.10: Predicted Salary VS True Salary for RF   Figure 5.11: Residual plot for RF

In Figure 5.10, compared to the first study's model, most blue points in the RF model are near to redline which means most predicted salaries in the RF model are closer to true salaries. Because the result of regression trees is based on averages from the training dataset's weekly salary, then there will be no negative values for the predicted salary. So, this model is furthermore ideal than the first model.

Figure 5.11 has nearly the same characteristics of the randomness and the normal distribution of residuals as the mixed-effect model's residual plot, the residuals of this RF model are also independent and normally distributed, so this RF regression model is also valid.

### 6.5. Underestimation of Overestimation of Players' Value by RF Predictive Model

#### 6.5.1. Method of Player's Value Estimation

The R-square value from comparing predicted salaries by RF and true salaries of all players in the dataset is good: $R^2 = 0.948$. Therefore, due to the accuracy of the predicted values, this study will judge the estimation of players by a more direct method rather than using the prediction interval. The study still evaluates the players by dividing them into 3 categories like the evaluation method in a mixed-effect study – Overestimation/Underestimation/Normal. The study directly compares the predicted salary with the true salary and measures the estimation based on their difference ratio, then the study hopes to find a threshold percentage to judge whether players belong to Overestimation/Underestimation/Normal.

Therefore, this study decides to calculate the Symmetric Absolute Percentage Error for each player's predicted salary and true salary, which can get a proportion to express the prediction error between true salary and predicted value. Moreover, the Symmetric Mean Absolute Percentage Error (SMAPE), which is the mean value of all players' Symmetric Absolute Percentage Error, will be regarded as a threshold to judge the estimation of players. the specific method is:

$$SMAPE = \frac{1}{n}\sum_{t=1}^{n}\frac{|P_t - A_t|}{(P_t + A_t)/2}; \quad P_t \text{ is predicted value}, A_t \text{ is actual value}$$

Compared to the Absolute Percentage Error and Mean and Mean Absolute Percentage Error (MAPE), both methods express the difference as the ratio, which is indeed obvious and understandable, however, one huge advantage for SMAPE is that the ratio values of SMAPE are limited in the interval between 0% and 200%, which can reduce the impact of the players with a huge difference between predicted salary and true salary on the total data's mean error.

The specific evaluation step is:
- Calculate the Symmetric Absolute Percentage Error for all players.
- For one of the players in the dataset:
- If the true salary is larger than the predicted salary and the Symmetric Absolute Percentage Error is larger than the SMAPE, then -> **Overestimation**.
- If the true salary is smaller or equal to the predicted salary and the Symmetric Absolute Percentage Error is larger than the SMAPE, then -> **Underestimation**.
- Otherwise -> **Normal**.

## 6.6. Conclusion of Player's Value Estimation

According to the calculation, the $SMAPE \approx 29.37\%$, and this value will be a threshold to judge the estimation of players.

Figure 5.12: The distribution of estimation of players by RF.

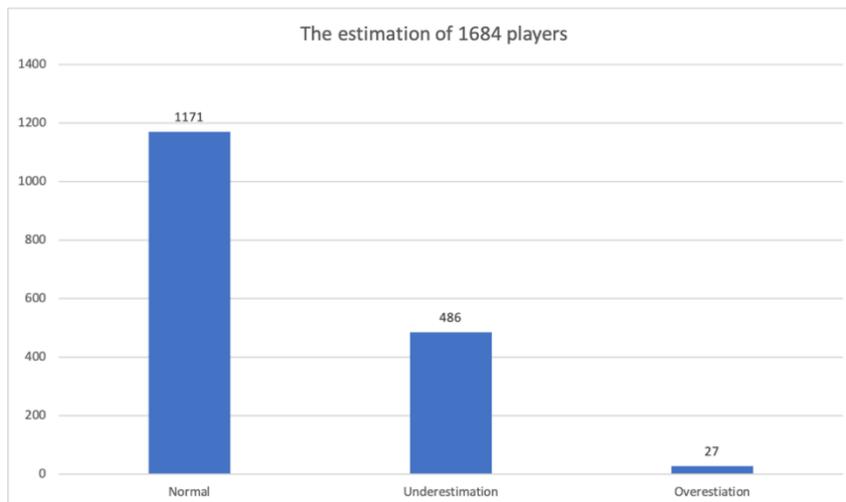

This part still displays the market value estimation of 1684 players in the dataset. The result of the judgement of 1684 players' estimation has shown in the figure 5.12 above: 1171 players (69.6%) are regarded as the normal estimation; 486 players (28.9%) are overestimation; 27 players (1.6%) are an underestimation.

Compared to the results of the mixed-effect model, normal estimated players have nearly the same proportion. But the underestimated players concluded from the RF are much larger than in the mixed-effect study and overestimated players are also much less than in the last study.

The estimation method of directly calculating the errors between predicted and actual values and setting the mean error as the threshold is normally stricter and more restrictive than using the prediction interval to judge, but still, nearly 69% of players will be regarded as the normal estimation of the players which is nearly the same result as the mixed-effect model, therefore, RF model's predicted salaries of players are closer to the actual salaries than a mixed-effect model.

## 6.2: RESEARCH CONTRIBUTION

This study contributes a lot to the football world even in the whole sports world in several ways. First, this study provides a new method to evaluate the value of players engaged in highly-commercialized sports – to use players' salaries as a proxy to reflect the value of the players. The second is to use a different type of ML model Ensemble method and choose the Random Forest algorithm to build the model, and this model may never be used by some other researchers to explore the market value of the players. The third is to provide two models to calculate the approximate salary of the players based on their selected features and even to evaluate whether players are overestimation or underestimation. The last one is to find some new factors that may be important to determine the market value and salary of players, especially for achievements of players, and the research not only proves that this factor has the hugest impact on the salary of players in all features but also finds an efficient way to combine all crucial tournaments into one index.

## CONCLUSION

When selecting the features and getting the data, this research fails to get the data expressing the popularity of players (e.g., Wikipedia page views, Reddit posts). Therefore, to make the research more comprehensive, the popularity data of players need to be obtained. Then based on the methods of this research, further study may build a more complete salary predictive model with higher accuracy and explore the relation between other related features of the players (like the popularity of players) and their salary or market value.

Moreover, the research only successfully gets the salary data of players in the top 5 leagues and does not get players' data in some other leagues or some secondary leagues, like leagues of Portugal and Netherlands or secondary leagues of the top 5 leagues. Therefore, this research can only be restrictive to the players and clubs in the top 5 leagues, and the training dataset is also not very large due to the limited number of players in the top 5 leagues.

Nevertheless, the fundamental objective of this research is to evaluate the market value of soccer players by using the salary of the players as the proxy and explore the importance of each feature to the salary of players. The evaluation of market value is not to calculate the specific market values of players because this index is unobservable. Hence the purpose of studies is to judge whether a soccer player is overestimated or underestimated based on comparing the true salary with the player's predicted salary from trained ML models. Also, from the statistical results of both studies, the research analyzes the significance and importance of factors that affect the salary of the players.

## Bibliography:

(n.d.). Retrieved from RDocumentation: https://www.rdocumentation.org/packages/merTools/versions/0.5.2/topics/predictInterval

Bates, D., Mächler, M., Bolker, B., & Walker, S. (2015, October 7). Fitting Linear Mixed-Effects Models Using lme4. *Journal of Statistical Software, 67*(1), pp. 1-48.

Beers, B. (2022, March 5). *What Is P-Value?* Retrieved from Investopedia: https://www.investopedia.com/terms/p/p-value.asp#:~:text=A%20p%2Dvalue%20is%20a,significance%20of%20the%20observed%20difference.

Brownlee, J. (2019, October 30). *Probabilistic Model Selection with AIC, BIC, and MDL*. Retrieved from Machine Learning Mastery: https://machinelearningmastery.com/probabilistic-model-selection-measures/

Carmichael, F., Forrest, D., & Simmons, R. (1999, April). THE LABOUR MARKET IN ASSOCIATION FOOTBALL: WHO GETS TRANSFERRED AND FOR HOW MUCH? *Bulletin of Economic Research, 51*(2), pp. 125-150.

*Cross-validation: evaluating estimator performance*. (n.d.). Retrieved from scikit-learn : https://scikit-learn.org/stable/modules/cross_validation.html#multimetric-cross-validation

Donalek, C. (2011, April). *Supervised and Unsupervised Learning*. Retrieved from https://sites.astro.caltech.edu/~george/aybi199/Donalek_classif1.pdf

*Dribble*. (n.d.). Retrieved from SPORTSDEFINITIONS.COM: https://www.sportsdefinitions.com/soccer/dribble/

*Ensemble methods*. (n.d.). Retrieved from scikit-learn: https://scikit-learn.org/stable/modules/ensemble.html

Franck, E., & Nüesch, S. (2012, January). TALENT AND/OR POPULARITY: WHAT DOES IT TAKE TO BE A SUPERSTAR? *Economic Inquiry, 50*(1), pp. 202-216.

Frick, B. (2006). Salary determination and the pay-performance relationship in professional soccer: evidence fron Germany. *Sports Economics After Fifty Years: Essays in Honour of Simon Rottenberg*, 125-146.

Frost, J. (n.d.). *Interpreting Correlation Coefficients*. Retrieved from Statistics By Jim: https://statisticsbyjim.com/basics/correlations/

Gohar, U. (2020, March 5). *How to use Residual Plots for regression model validation?* Retrieved from Towards Data Science: https://towardsdatascience.com/how-to-use-residual-plots-for-regression-model-validation-c3c70e8ab378

Hastie, T., Tibshirani, R., & Friedman, J. (2016). *The Elements of Statistical Learning: Data Mining, Inference, and Prediction.* Springer.

He, M., Cachucho, R., & Knobbe, A. (2015, September). Football player's performance and market value.

Herm, S., Callsen-Bracker, H.-M., & Kreis, H. (2014, November). When the crowd evaluates soccer players' market values: Accuracy and evaluation attributes of an online community. *Sport Management Review, 17*(4), pp. 484-492.

Kaplan, T. (2010, July 8). *When It Comes to Stats, Soccer Seldom Counts.* Retrieved from The New York Times: https://www.nytimes.com/2010/07/09/sports/soccer/09soccerstats.html

Lucifora, C., & Simmons, R. (2003, February). Superstar Effects in Sport Evidence From Italian Soccer. *Journal of Sports Economics, 4*(1), pp. 35-55.

Müller, O., Simons, A., & Weinmann, M. (2017, May). European Journal of Operational Research. *European Journal of Operational Research, 263*(2), pp. 611-624.

Nagpal, A. (2017, October 17). *Decision Tree Ensembles- Bagging and Boosting*. Retrieved from Towards Data Science: https://towardsdatascience.com/decision-tree-ensembles-bagging-and-boosting-266a8ba60fd9

Nakagawa, S., Johnson, P. C., & Schielzeth, H. (2017, September ). The coefficient of determination R2 and intra-class correlation coefficient from generalized linear mixed-effects models revisited and expanded. *JOURNAL OF THE ROYAL SOCIETY INTERFEACE*.

*OPTA EVENT DEFINITIONS*. (n.d.). Retrieved from STATS PERFORM: https://www.statsperform.com/opta-event-definitions/#:~:text=A%20shot%20on%20target%20is%20defined%20as%20any%20goal%20attempt%20that%3A&text=Goes%20into%20the%20net%20regardless%20of%20intent.&text=Is%20a%20clear%20attempt%20to,goal%20(last%20line%20bloc

Roback, P., & Legler, J. (2020). *Beyond Multiple Linear Regression.* Chapman and Hall/CRC.

Shrikanth, S. (n.d.). *A Language, not a Letter: Learning Statistics in R.* Retrieved from Chapter 17: Mixed Effects Modeling: https://ademos.people.uic.edu/Chapter17.html

*sklearn.ensemble.RandomForestRegressor*. (n.d.). Retrieved from scikit-learn: https://scikit-learn.org/stable/modules/generated/sklearn.ensemble.RandomForestRegressor.html

Snijders, T. A. (2011). *Multilevel Analysis: An Introduction To Basic And Advanced Multilevel Modeling.* SAGE Publications Ltd.

*Soccer Blade*. (n.d.). Retrieved from https://soccerblade.com/blocking-in-soccer/

*Soccer Intercept*. (n.d.). Retrieved from rookie road: https://www.rookieroad.com/soccer/intercept/

*Soccer Passing Rules*. (n.d.). Retrieved from rookie road: https://www.rookieroad.com/soccer/basics/passing/

*Soccer Red Cards*. (n.d.). Retrieved from rookie road: https://www.rookieroad.com/soccer/red-cards/

*Soccer Shooting*. (n.d.). Retrieved from rookie road:
    https://www.rookieroad.com/soccer/basics/shooting/

*Soccer Tackle*. (n.d.). Retrieved from rookie road:
    https://www.rookieroad.com/soccer/basics/tackling/

*Soccer Yellow Card*. (n.d.). Retrieved from rookie road: https://www.rookieroad.com/soccer/yellow-card/

SPORTYTELL. (2021, January 16). *Top-10 Most Popular Sports In The World 2021*. Retrieved from
    https://sportytell.com/sports/most-popular-sports-world/

Trequattrini, R., Lombardi, R., & Nappo, F. (2012, October). The evaluation of the economic value of
    long lasting professional football player performance rights. *WSEAS TRANSACTIONS on
    BUSINESS and ECONOMICS, 9*(4), pp. 199-218.

UEFA. (2020). *FINANCIAL REPORT 2019/20.*

UEFA. (2021). Retrieved from UEFA CLUB COEFFICIENTS:
    https://www.uefa.com/nationalassociations/uefarankings/country/#/yr/2021

UEFA. (2021). *The European Club Footballing Landscape.*

*Vocabulary: Football*. (2011). Retrieved from BBC World Service:
    https://www.bbc.co.uk/worldservice/learningenglish/grammar/vocabulary/football.shtml

Zhou, Z.-H. (2012). *Ensemble Methods Foundations and Algorithms.* New York: Chapman and
    Hall/CRC.## Appendix

**GitHub Link**: https://github.com/leolee20/Final-year-project

Note: The GitHub repository does not include any codes about processing the original datasets crawled from websites https://www.capology.com and https://fbref.com/en/. It includes the final processed data based on Chapter 3 and the two trained models based on Chapter 4, and it also contains the codes for building the models, training the models, and using the model to do predictions. Then the GitHub repository also includes the codes for analyzing the results of models, and most importantly, it contains the file to summarize both models' predicted salaries, estimation of their value, and the comparison of both models' estimation results.

30